\documentclass{bmvc2k}
\usepackage{amsfonts}
\usepackage{tabularx}
\usepackage{multirow}
\usepackage{subfigure}
\graphicspath{{images/}}
\usepackage{tabularx}
\usepackage{float}

\title{Feature Pyramid Encoding Network for Real-time Semantic Segmentation}

\addauthor{Mengyu Liu}{mengyu.liu@manchester.ac.uk}{1}
\addauthor{Hujun Yin}{hujun.yin@manchester.ac.uk}{1}

\addinstitution{
 School of Electrical and Electronic Engineering\\
 The University of Manchester\\
 Manchester, UK
}

\runninghead{LIU AND YIN}{FPENet for Real-time Semantic Segmentation}

\def\etal{\emph{et al}\bmvaOneDot}

\begin{document}

\maketitle

\begin{abstract}
Although current deep learning methods have achieved impressive results for semantic segmentation, they incur high computational costs and have a huge number of parameters. For real-time applications, inference speed and memory usage are two important factors. To address the challenge, we propose a lightweight feature pyramid encoding network (FPENet) to make a good trade-off between accuracy and speed. Specifically, we use a feature pyramid encoding block to encode multi-scale contextual features with depthwise dilated convolutions in all stages of the encoder. A mutual embedding upsample module is introduced in the decoder to aggregate the high-level semantic features and low-level spatial details efficiently. The proposed network outperforms existing real-time methods with fewer parameters and improved inference speed on the Cityscapes and CamVid benchmark datasets. Specifically, FPENet achieves 68.0\% mean IoU on the Cityscapes test set with only 0.4M parameters and 102 FPS speed on an NVIDIA TITAN V GPU.      
\end{abstract}

\section{Introduction}
\label{sec:intro}
Semantic segmentation has become one of the popular research areas with the recent success of deep convolutional neural networks (CNNs). It aims to assign a particular class to each pixel of an image, and can be applied to many applications from self-driving vehicles to medical image diagnostics. Most of the state-of-the-art semantic segmentation models are based on the fully convolutional network (FCN) \cite{long2015} to provide end-to-end dense classification in images, and some employ conditional random fields (CRFs) \cite{krahenbuhl2011} as a post-processing method to refine the boundaries of segmentation results. Most of the high performing methods often have a large number of parameters due to their deep and wide architectures. For example, PSPNet \cite{zhao2017} has 65.7 million parameters and DeepLabV3+ \cite{chen2018encoder} contains 54.6 million parameters. Besides, these methods require huge computational resources and take a long time to process an image even on modern GPUs. However, reality applications of semantic segmentation usually require real-time inference and low memory footprint.

To address the above problem, several real-time semantic segmentation methods \cite{zhao2018icnet, paszke2016, poudel2018} have been proposed to make a trade-off between accuracy and speed. Some methods take downsampled input images to reduce the computation complexity and fuse features at different levels \cite{zhao2018icnet, poudel2018}, while others prune redundant channels to reduce the number of parameters \cite{paszke2016}. These methods have achieved faster inference speed at the cost of lower accuracy on benchmarks \cite{cordts2016, everingham2010}. Features extracted from downsampled images lack spatial details, and pruned shallow networks are weak in encoding contextual information with small receptive fields.   

Most of the semantic segmentation models employ the U-shape architecture \cite{ronneberger2015}, which is composed of a deep encoder to extract features and a decoder to fuse the extracted features at different levels for final pixel-level classification. Most of the real-time segmentation models contain light decoders, consisting of few convolutional layers and bilinear upsampling to recover resolution \cite{wu2018, paszke2016}. These simple decoders reduce number of parameters and increase speed, but the fine information is lost, leading to coarse segmentation, especially at boundaries. Some of the high performing methods employ complicated decoders to fuse high-level features with low-level features \cite{peng2017, jegou2017, zhang2018}, hence spatial information can be preserved to produce fine segmentation in this way. However, these methods have increased computational complexity, leading to low efficiency.

Based on these observations, a feature pyramid encoding network (FPENet) for real-time semantic segmentation is proposed. It is a lightweight U-shape model consisting of an encoder and a decoder. In the encoder, the feature pyramid encoding (FPE) block combines a pyramid of dilated convolutions with depth-separable inverted bottleneck block \cite{sandler2018}. Groups of depthwise dilated convolutions of different rates are employed in the FPE block to perform as a spatial pyramid and reduce computational complexity. Encoding multi-scale features with different sizes of receptive fields has been proven helpful for semantic segmentation \cite{zhao2017, chen2018encoder, yang2018}. Instead of placing the spatial pyramid module at the end of the network, we employ it in each block to model spatial dependency and learn representations from feature maps at different levels. Depth-separable convolutions \cite{howard2017} are combined with dilation convolutions in the FPE block to reduce the number of parameters and inference time. For the decoder, in order to aggregate features of different levels efficiently, we propose a mutual embedding upsample (MEU) module, which uses global contextual concepts from high-level features to guide low-level features and embeds local spatial information from low-level features into high-level features simultaneously.

In summary, the main contributions are as follows. 

(\romannumeral1) A feature pyramid encoding block is proposed to encode multi-scale features and reduce computational complexity with groups of pyramid depthwise dilated convolutions. 

(\romannumeral2) A mutual embedding upsample module is introduced to aggregate the high-level and low-level features. 

(\romannumeral3) Significant improvements are obtained on the Cityscapes \cite{cordts2016} and CamVid \cite{brostow2008} benchmarks, with similar number of  parameters but much faster inference speed compared to the existing segmentation methods.

\section{Related Work}
First we review recent developments in real-time semantic segmentation. Multiple studies have explored the impact of encoding multi-level contextual features with large receptive fields. Finally, we summarize the recent research focused on feature aggregation. 

\textbf{Real-time segmentation algorithms:} Real-time segmentation algorithms are required to make a trade-off between accuracy and speed, and these models are expected to be lightweight. In ICNet \cite{zhao2018icnet} and ContextNet \cite{poudel2018}, multi-scale images were employed as inputs of cascaded networks to extract features. Downsampled images were applied to deep branches while large images were applied to shallow branches in these two models to reduce computational complexity. ENet \cite{paszke2016} discards the last stage of the network and reduces the number of downsampling times to shrink the model. Mehta \etal. proposed the ESPNet \cite{mehta2018}, where efficient pyramid modules were utilized to extract multi-scale features. BiSeNet \cite{yu2018} extracts high-level semantic features and low-level spatial information independently with two paths. CGNet \cite{wu2018} learns the joint representations of local features and their surrounding context, and utilizes global context attention to refine the joint features.   

\textbf{Multi-level contextual features:} Encoding contextual features at multiple levels helps achieve good results in semantic segmentation due to multiple scales of objects and spatial dependency. Zhao \etal. showed that global contextual features were beneficial for semantic segmentation, and proposed the PSPNet \cite{zhao2017}, which applied a multi-scale spatial pooling module at the end of the model to exploit multi-level contextual features by pooling operations. In \cite{chen2018deeplab}, an atrous spatial pyramid pooling (ASPP) module was proposed to model semantic contextual information. ASPP contains several parallel atrous (dilated) convolutions of different rates, and multi-level contextual features are encoded simultaneously. Yang \etal. improved the ASPP module by a DenseASPP block \cite{yang2018}, where the dilated convolutions were connected in a dense way to generate densely sampled features. In the pyramid attention network (PAN) \cite{li2018}, the spatial pyramid pooling was combined with attention to generate precise pixel-level attention for high-level contextual features. In Res2Net \cite{gao2019res2net}, a group of $3\times3$ filters in residual block was replaced with smaller groups of filters to extract contextual information simultaneously.          

\textbf{Feature aggregation:} Because of the repeated downsampling layers in CNNs, directly upsampling the final score map to the original resolution would lead to coarse results and loss of fine details. FCN adopts skip connections which combine the coarse and fine predictions to reconstruct dense feature maps. Ronneberger \etal. proposed a U-shape network \cite{ronneberger2015}, which was composed of an encoder and a symmetric decoder, and long skip connections were introduced to link these two parts. Peng \etal. utilized boundary refinement modules in the decoder to enhance feature aggregation ability \cite{peng2017}. Li \etal. proposed a global attention upsample module in the decoder to extract global context of high-level features as guidance to weight low-level feature information \cite{li2018}. In \cite{zhang2018}, the effectiveness of feature fusion at different levels was explored, and deeply supervised training and semantic supervision were applied to low-level features to introduce more semantic concept.   

\section{Methods}
We here present the feature pyramid encoding (FPE) block and the mutual embedding upsample (MEU) module in detail. The complete network architecture is then described. 

\begin{figure*}[h!]
	\centering    	
	\subfigure[Depth-separable inverted bottleneck block]{
		\includegraphics[height=0.47\linewidth]{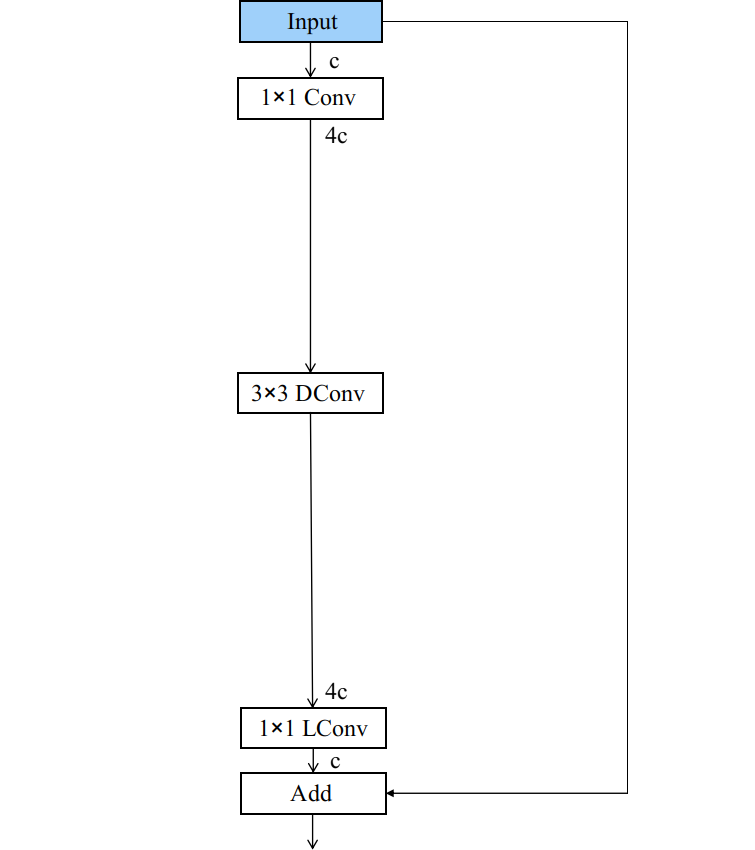}
		\label{fca}}	
	\subfigure[FPE block]{
		\includegraphics[height=0.47\linewidth]{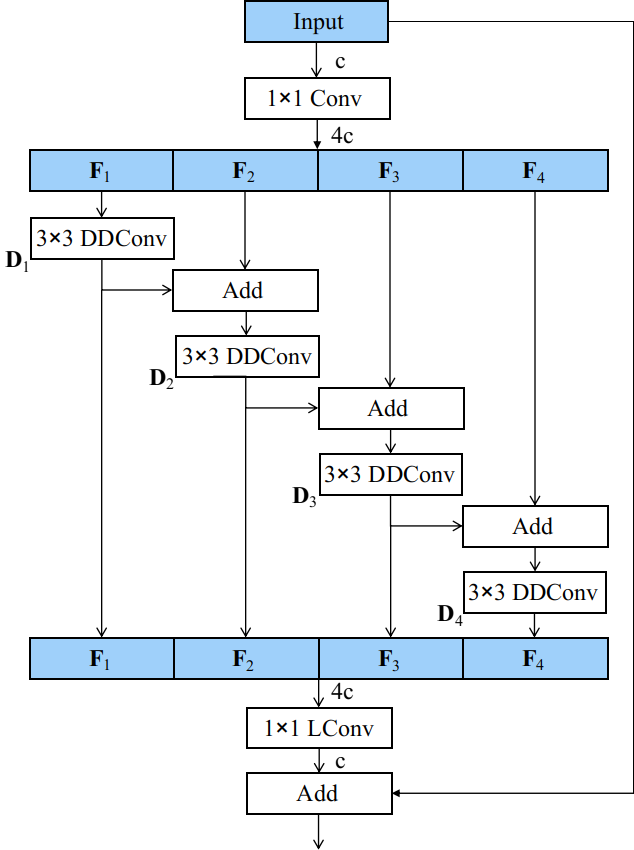}
		\label{fpeb}}	
	\caption{Structures of (a) depth-separable inverted bottleneck block and (b) FPE block. The expansion ratio is 4, and dilation rates in FPE block are 1, 2, 4, 8, respectively. DConv: depthwise convolution. LConv: linear convolution. DDConv: depthwise dilated convolution. c: the number of input channels.}
	\label{fpe}
\end{figure*}
\subsection{FPE Block}

Many approaches \cite{chen2018deeplab, zhao2017, li2018} encode multi-scale features with ASPP or a pyramid pooling module at the end of the model to increase receptive field, while others \cite{mehta2018, mehta2019espnetv2, wu2018} adopt parallel dilated convolutions with different rates in each stage of the network to combine local information with surrounding context. Encoding multi-scale features simultaneously can yield better performance of semantic segmentation. We combine dilated convolutions with inverted bottleneck structure to perform pyramid encoding in each block of the network. 

The FPE block is based on the depth-separable inverted bottleneck block \cite{sandler2018} and is composed of a $1\times1$ expansion convolutional layer, groups of $3\times3$ depthwise convolutions and a final $1\times1$ pointwise convolution, and residual connection is employed where the number of input channels is equal to the number of output channels. The number of channels is expanded $4$ times by the first $1\times1$ convolution and squeezed $4$ times by the final $1\times1$ convolution. Depthwise convolution splits the input into $N$ ($N$ is the number of input channels) groups, then an independent single-channel convolutional filter is applied to each channel. After this, a pointwise convolution is used to fuse these outputs linearly. The combination of depthwise convolution and pointwise convolution is extremely efficient, as it reduces by around 9 times the computational cost compared to the standard convolution \cite{howard2017}. 

Figure \ref{FPE} shows the differences between the depth-separable inverted bottleneck block and the proposed FPE block. For an input feature map of size $w \times h \times c$ where $w$, $h$ are the spatial width and height of the feature map, respectively, and $c$ is the number of input channels, the FPE block first expands the number of channels from $c$ to $4c$ using $1\times1$ convolution. Similar to the Res2Net module \cite{gao2019res2net}, the output feature map is split into 4 subsets of $c$ channels, denoted by $\textbf{F}_{i}$, $i \in \left\lbrace 1,...,4\right\rbrace $. And then each subset is processed by a group of $3\times3$ depthwise dilated filters $\textbf{D}_i$. The output of $\textbf{D}_{i}$ is added to the following subset $\textbf{F}_{i+1}$, and then processed by $\textbf{D}_{i+1}$. The outputs of these parallel branches are concatenated and then fused by the final $1\times1$ linear convolution to reduce to $c$ channels.     
     
The pyramid encoding mechanism is performed by these four parallel depthwise dilated convolutions, and the dilation rate of $\textbf{D}_{i}$ is $2^{i-1}$. Dilated convolutions enlarge the size of receptive field by inserting zeros between weights of convolutional kernels without increasing parameters. For a normal bottleneck block, the receptive field is only $3\times3$, while the receptive field of FPE block is up to $17\times17$. Branch $\textbf{D}_{i}$ processes all the features extracted from the previous branches to enhance information flow, and the number of pixels participate in computation increases with the dilation rate. This structure can be considered as four spatial pyramid encoding modules, where the dilation rate increases one by one, and contextual features are encoded under four scales. The final output of FPE block is a feature map generated by multi-scale features, which carries local and surrounding contextual information.

\begin{figure}[t!]
	\centering    
	\subfigure[MEU]{
		\includegraphics[width=0.5\textwidth]{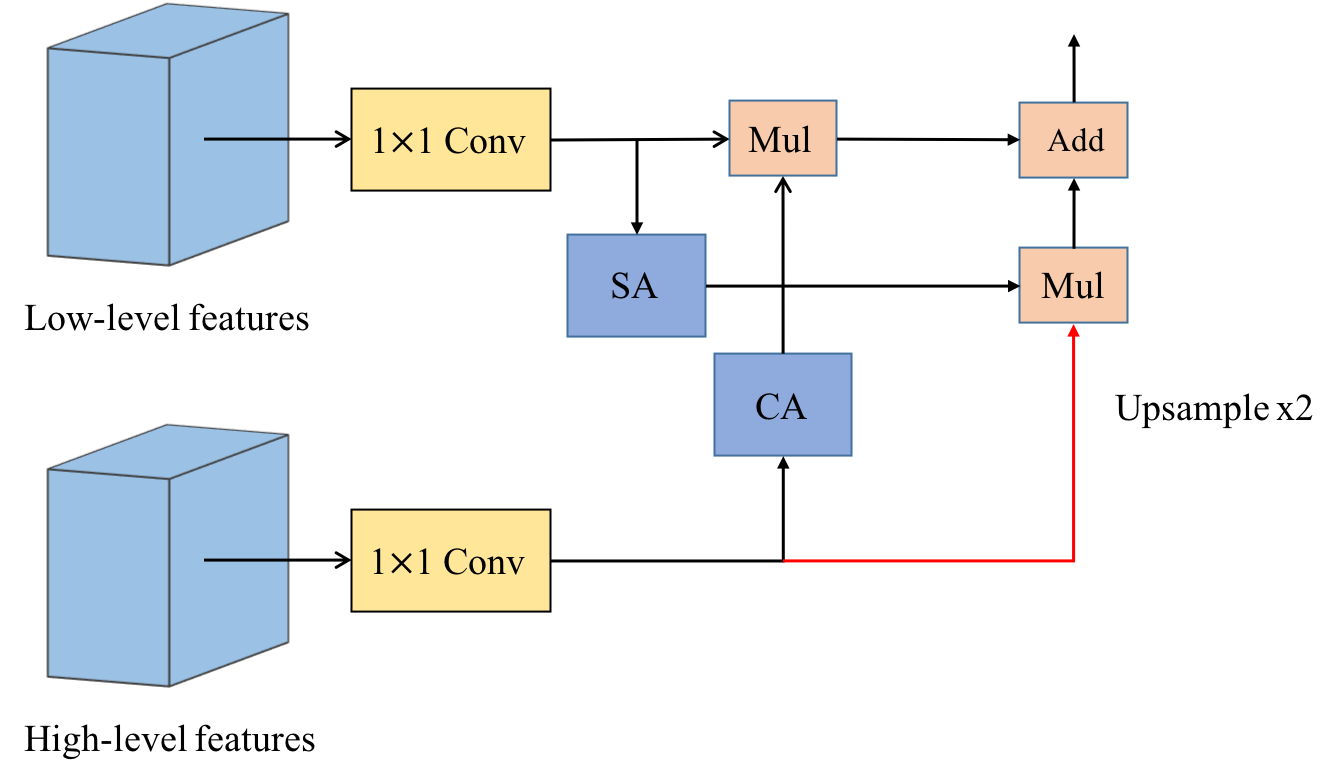}
		\label{meu}}		
		\begin{minipage}[b]{0.45\textwidth}
		\centering
		\subfigure[SA]{
		\includegraphics[width=1\textwidth]{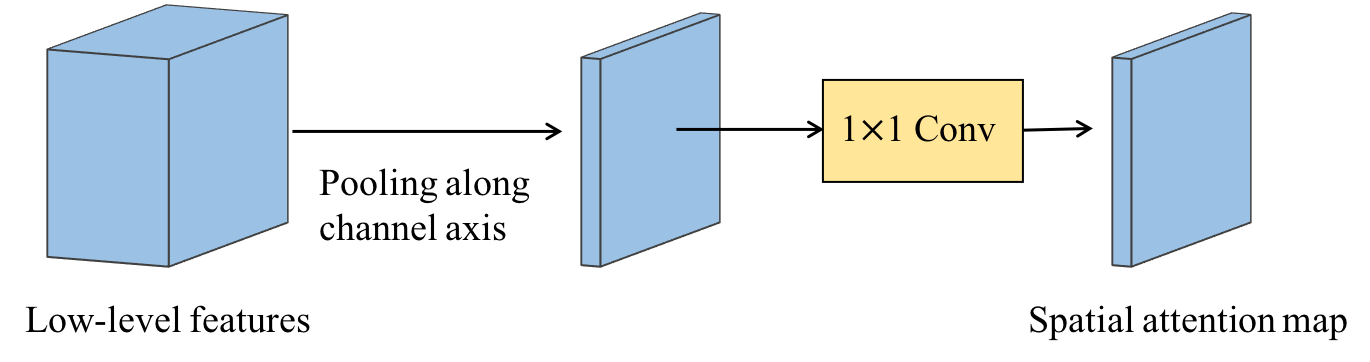}
		\label{sa}}
	\subfigure[CA]{
		\includegraphics[width=1\textwidth]{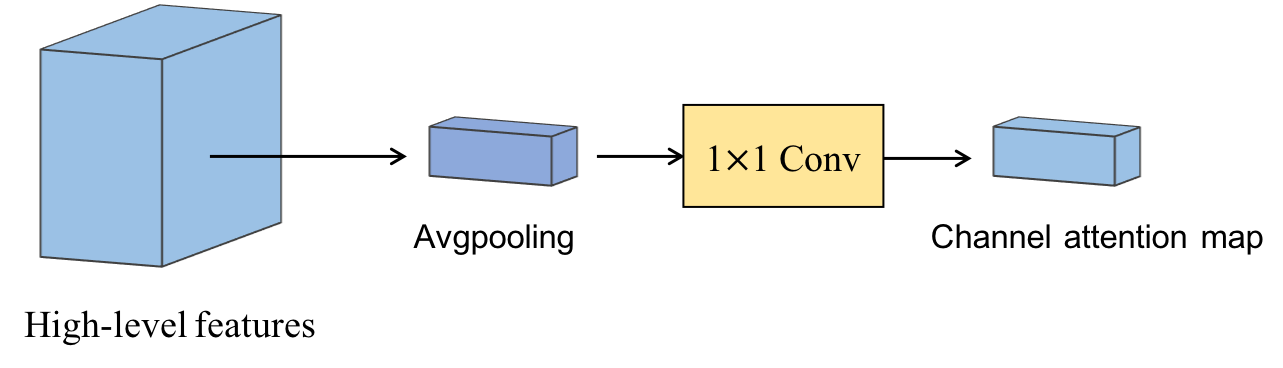}
		\label{ca}}
	\end{minipage}
	\caption{(a) Structure of MEU module. SA: spatial attention block. CA: channel attention block. (b) Spatial attention block. (c) Channel attention block.}
	\label{MEU}
\end{figure}

\begin{figure*}[t!]
	\centering    
	\includegraphics[width=0.67\textwidth]{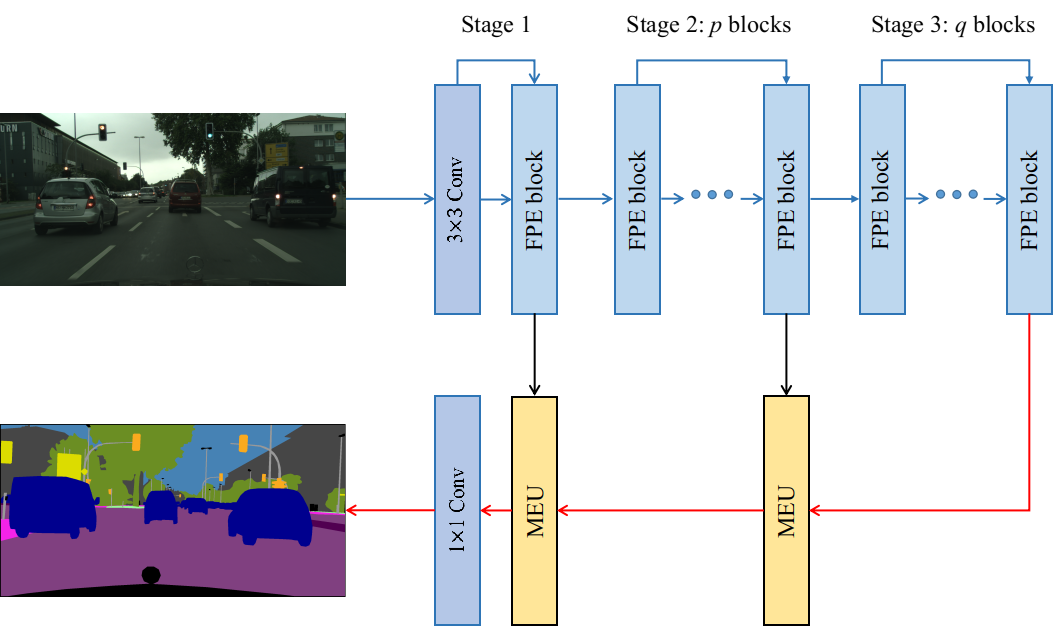}		
	\caption{Architecture of FPENet.}
	\label{fpenet}
\end{figure*}

\subsection{MEU Module}
In U-shape models, decoder is designed to aggregate features extracted at different levels to recover the resolution. Many methods \cite{chen2018encoder, zhao2017, zhang2018} use bilinear upsampling or several simple convolutions as a naive decoder. These naive decoders only consider high-level semantic concepts and ignore low-level spatial details leading to coarse segmentation. While other approaches \cite{peng2017, jegou2017, lin2017refinenet} adopt complicated modules in decoders to aggregate features from different stages and utilize low-level features to refine boundaries. However, these well-designed decoders are time-consuming.

High-level features contain contextual information while low-level features are rich in spatial details. This makes feature aggregation difficult. Zhang \etal. showed that introducing more contextual information into low-level features or embedding more spatial details into high-level features can enhance feature fusion \cite{zhang2018}. PAN \cite{li2018} adopts a global attention upsample module to squeeze high-level context and embeds it into low-level features as a guidance. We consider that low-level features containing rich spatial information can also be embedded into high-level features as a guidance.       

The MEU module consists of two attention blocks as depicted in Figure \ref{MEU}. First, two $1\times1$ convolutions are performed on the high-level and low-level features, respectively. Next, the high-level features from the channel attention block go through a global average pooling operation, a $1\times1$ convolution and a ReLU operator, and then are multiplied by the low-level features. While in the spatial attention block, low-level features are first squeezed by an average pooling operation along the channel axis, next a $1\times1$ convolution and a ReLU non-linearity are applied to generate a single-channel attention map, which is then multiplied by the upsampled high-level features. Finally, these two weighted features are fused by element-wise addition.    

The spatial attention map generated from low-level features corresponds to the importance of each pixel. It focuses on localizing the objects and refining the boundaries with spatial details. While the squeezed channel attention map generated from high-level features reflects the importance of each channel. It focuses on the global context to provide content information. The MEU module extracts these two kinds of attention maps and efficiently embeds semantic concepts and spatial details to low-level and high-level features.

\begin{table}[t!]
	\begin{center}
		\begin{tabular}{ l  c  c c }
			\hline
			Name                    & Operator             & Channel & Output size    \\ \hline\hline
			\multirow{2}{*}{stage1} & $3\times3$ Conv      & 16      & $512\times256$ \\
			                        & FPE $(k=1) \times1$  & 16      & $512\times256$ \\ \hline
			stage2                  & FPE $(k=4) \times p$ & 32      & $256\times128$ \\ \hline
			stage3                  & FPE $(k=4) \times q$ & 64      & $128\times64$  \\ \hline
			decoder2                & MEU                  & 64      & $256\times128$ \\ \hline
			decoder1                & MEU                  & 32      & $512\times256$ \\ \hline
			final                   & $1\times1$ Conv      & $C$     & $512\times256$ \\ \hline
		\end{tabular}
	\end{center}
	\caption{Architecture details of FPENet. Input size is $3\times1024\times512$. $k$ is the expansion ratio of FPE block. $C$ is the number of classes.}
	\label{FPE}
\end{table}

\subsection{Network Architecture}        
The entire network architecture is shown in Figure \ref{fpenet}. Based on the above discussion, we have designed this lightweight encoder-decoder model with FPE blocks and MEU modules. In order to preserve spatial information and reduce number of parameters, the total downsampling rate is 8. The detailed structure of the proposed model is shown in Table \ref{FPE}. 

We employ FPE blocks in the encoder except the first layer, and the number of channels in each stage is 16, 32, 64, respectively. In stages 2 and 3, we employ $p$ and $q$ FPE blocks respectively, and the stride of $3\times3$ depthwise dilated convolutions is set to 2 in the first blocks to downsample feature maps. All expansion ratios of FPE blocks are set to 4 to perform pyramid encoding except for the first, a normal bottleneck block. We add long skip connections in stages 2 and 3, the inputs of these two stages are combined from the outputs of the first and last blocks of their preceding stages. These skip connections encourage signal propagation and perform as an implicit deep supervision, cause earlier layers to connect to the deepest layer to receive supervision from different stages of the decoder. For the decoder, two MEU modules are used to aggregate features from each stage and recover the resolution step by step. Finally, a $1\times1$ convolutional layer is applied as the pixel-level classifier.

\section{Experiments}

\subsection{Implementation Protocol}
We conducted all the experiments using PyTorch \cite{paszke2017} with CUDA 10.0 and cuDNN back-ends. Adam algorithm \cite{kingma2014} with batch size 8 and weight decay 0.0001 were used to train the networks from scratch without any pre-training on any large datasets. The ``poly'' learning rate policy \cite{chen2018deeplab} was employed:

\begin{equation}
lr=init\;lr \times (1-\frac{epoch}{max\_epoch})^{power}
\label{poly}
\end{equation}
where $epoch$ is the current number of epoch, $power$ is 0.9 and initial learning rate was set to 0.0005. We employed the zero-mean normalization, random horizontal flip, random rotation between -10 and 10 degree and random scaling between 0.5 and 1.75 for data augmentation. The networks were trained for 400 epochs on the Cityscapes and 300 epochs on the CamVid. For training and test on the Cityscapes dataset, we downsampled the input images by two and recovered the segmentation results to original resolution using bilinear upsampling. For the CamVid dataset, images were trained and evaluated at the original resolution. Accuracy was measured using the mean Intersection-over-Union (mIoU) metric. The mean of cross-entropy error over all pixels was applied as the loss.  

\subsection{Ablation Studies}
The Cityscapes is an urban street scene dataset for semantic understanding. It contains 5000 fine annotated images, divided into three sets, 2975 for training, 500 for validation and 1525 for test. Furthermore, 20000 coarsely annotated images are provided for training. All images are of resolution, $2048\times1024$, and all pixels are annotated to 19 classes. In our experiments, only the fine annotated images were used for training the networks. In these ablation studies, we evaluated our networks on the validation set of Cityscapes to investigate the effect of each component in FPENet.

\textbf{Ablation on pyramid encoding structure:} We adopted three schemes to evaluate the effect of the pyramid encoding structure by changing the number of branches in the FPE block to 1, 2 and 4. When the number of branches is 1, the FPE block is equal to the normal bottleneck block. The expansion ratios were the same in these three schemes to keep number of parameters same, $p$ and $q$ were set to 3 and 7, respectively. Naive bilinear upsampling was employed as the decoder in these schemes. Results are shown in Table \ref{pyramid}, showing that the pyramid encoding structure gave better result, and two schemes improved the segmentation quality by 3.5\% and 6.6\%, respectively. These statistically significant improvements indicate that the pyramid encoding structure is beneficial for segmentation task as multi-scale contextual features are encoded efficiently without introducing new parameters.       

\begin{table}[htb]
			\begin{minipage}{0.47\linewidth}
				\centering		
				\begin{tabularx}{5.7cm}{p{1.5cm} | p{1.4cm}<{\centering} | X<{\centering}}
					\hline
					Name      & \#Branches & mIoU (\%) \\ \hline\hline
					FPE\_P3Q7 & 1          & 55.9      \\
					FPE\_P3Q7 & 2          & 59.4      \\
					FPE\_P3Q7 & 4          & 62.5      \\ \hline
				\end{tabularx}
			\caption{Results of FPE encoder with different number of branches.}
			\label{pyramid}
			\end{minipage}\quad
    \begin{minipage}{0.47\linewidth}
    	\centering		
	\begin{tabularx}{6.2cm}{p{1.5cm} | p{1.9cm}<{\centering} | X<{\centering}}
		\hline
		Name      & Dilation rates & mIoU (\%) \\ \hline\hline
		FPE\_P3Q7 & 1, 2, 3, 4     & 61.7      \\
		FPE\_P3Q7 & 1, 2, 4, 8     & 62.5      \\ \hline
	\end{tabularx}
	\caption{Results of FPE encoder with different combinations of dilation rates.}
	\label{rate}
\end{minipage}
\end{table}

\textbf{Ablation on dilation rates:} We designed two kinds of FPE blocks with different combinations of dilation rates and used them to build the encoder, one with dilation rates of 1, 2, 3, 4, while the another with 1, 2, 4, 8. As shown in Table \ref{rate}, the model with larger dilation rates in FPE block achieved better result. The range of receptive field of the former FPE block was from $3\times3$ to $9\times9$, while the latter $3\times3$ to $17\times17$. Larger receptive field can encode more surrounding features and learn better multi-scale representations.      

\textbf{Ablation on addition between branches:} In FPE blocks, we added the output of one branch to the input of following branch. As shown in Table \ref{add}, the addition operations between adjacent branches improved the accuracy from 62.5\% to 63.0\%. This improvement comes from the addition operations which change the independent branches to a cascaded pyramid module, so larger dilated convolutions perform on the features extracted by smaller dilated convolutions. The number of pixels convoluted by large kernels is also increased, this structure is similar to the DenseASPP module in \cite{yang2018}.

\begin{table}[t!]
	\begin{minipage}{0.46\linewidth}
	\centering
	\begin{tabular}{ l  c  c  c c}
		\hline
		$p$ & $q$ & \#Params & FLOPs & mIoU (\%) \\ \hline\hline
		3   & 5   & 233K     & 3.77G & 59.5      \\ \hline
		3   & 7   & 305K     & 4.37G & 64.1      \\
		5   & 7   & 325K     & 5.04G & 64.3      \\ \hline
		3   & 9   & 378K     & 4.98G & 65.5      \\
		5   & 9   & 398K     & 5.64G & 65.6      \\ \hline
		3   & 11  & 450K     & 5.58G & 65.8      \\ \hline
	\end{tabular}
	\caption{Results of FPENet with different depths, number of parameters and FLOPS are estimated on $1024\times512$ input.}
	\label{depth}
\end{minipage}\quad\quad
\begin{minipage}{0.47\linewidth}
	\centering	
	\begin{minipage}{0.9\linewidth}
	\begin{tabular}{  c  c  c}
		\hline
		Addition & Long skip & mIoU (\%) \\ \hline\hline
		         &           & 62.5      \\
		$\surd$  &           & 63.0      \\
		$\surd$  & $\surd$   & 64.1      \\ \hline
	\end{tabular}
\caption{Results of FPE encoder with different settings. $p = 3$, $q = 7$.}
\label{add}
\end{minipage}

\begin{minipage}{0.9\linewidth}
	\centering	
	\begin{tabular}{ c  c  c  c}
		\hline
		MEU & CA      & SA      & mIoU (\%) \\ \hline\hline
		w/o & ---     & ---     & 65.5      \\
		w   & $\surd$ &         & 66.5      \\
		w   & $\surd$ & $\surd$ & 67.2      \\ \hline
	\end{tabular}
	\caption{Results of MEU module with different components. $p = 3$, $q = 9$.}
	\label{decoder}
\end{minipage}
\end{minipage}	
\end{table}

\textbf{Ablation on long skip connection:} Long skip connections were employed in stages 2 and 3 in FPENet to combine the outputs of the first and final blocks. Accuracy was improved by 1.1\% as shown in Table \ref{add}. Intuitively, long skip connections apply implicit supervision to earlier layers and increase flow of information.

\textbf{Ablation on encoder depth:} We used different numbers of blocks in stages 2 and 3 to change the depth of the encoder. The numbers of parameters, FLOPs and accuracies of different configurations are shown in Table \ref{depth}. We can see that the value of $q$ has more impact on accuracy than $p$, indicating that stacking more FPE blocks increases receptive field in stage 3 and achieves better results. However, raising $q$ from 9 to 11, the improvement became minor, this may due to that the large receptive field in stage 3 is beyond the size of feature maps, and efficient features can not by extracted. Therefore, to make a trade-off between accuracy and computational complexity, we set $p$ to 3 and $q$ to 9 in the final architecture. 

\textbf{Ablation on decoder:} Since the FPE blocks extract features at different stages, MEU modules were used to aggregate these features to provide dense pixel-level prediction. We first evaluated the MEU module with only channel attention block, then we used channel and spatial attention together in MEU module to test the performance. As shown in Table \ref{decoder}, the channel and spatial attention blocks both improved the accuracy, indicating that embedding semantic concepts into low-level features and spatial details into high-level features with the MEU module lead to better results.   

\subsection{Cityscapes}
Based on the ablation studies, we combined the FPE blocks and MEU modules to build the complete network and experimented it on the Cityscapes dataset. First, we conducted experiments to estimate the inference speed at different resolutions for comparison with other methods. All experiments were conducted on an NVIDIA TITAN V GPU, using PyTorch framework with CUDA 10.0 and cuDNN 7.4, and each network was randomly initialized and evaluated for 100 times. The results and corresponding input sizes are shown in Table \ref{city}. Next, we trained FPENet with only fine annotated images of Cityscapes and accuracies on the test set are shown in Table \ref{city}. For a fair comparison, we did not employ multi-scale or multi-crop test. 

\begin{figure}[t!]
	\centering    
	\subfigure[Image]{
		\begin{minipage}[t]{0.19\textwidth}
			\centering
			\includegraphics[width=1\textwidth]{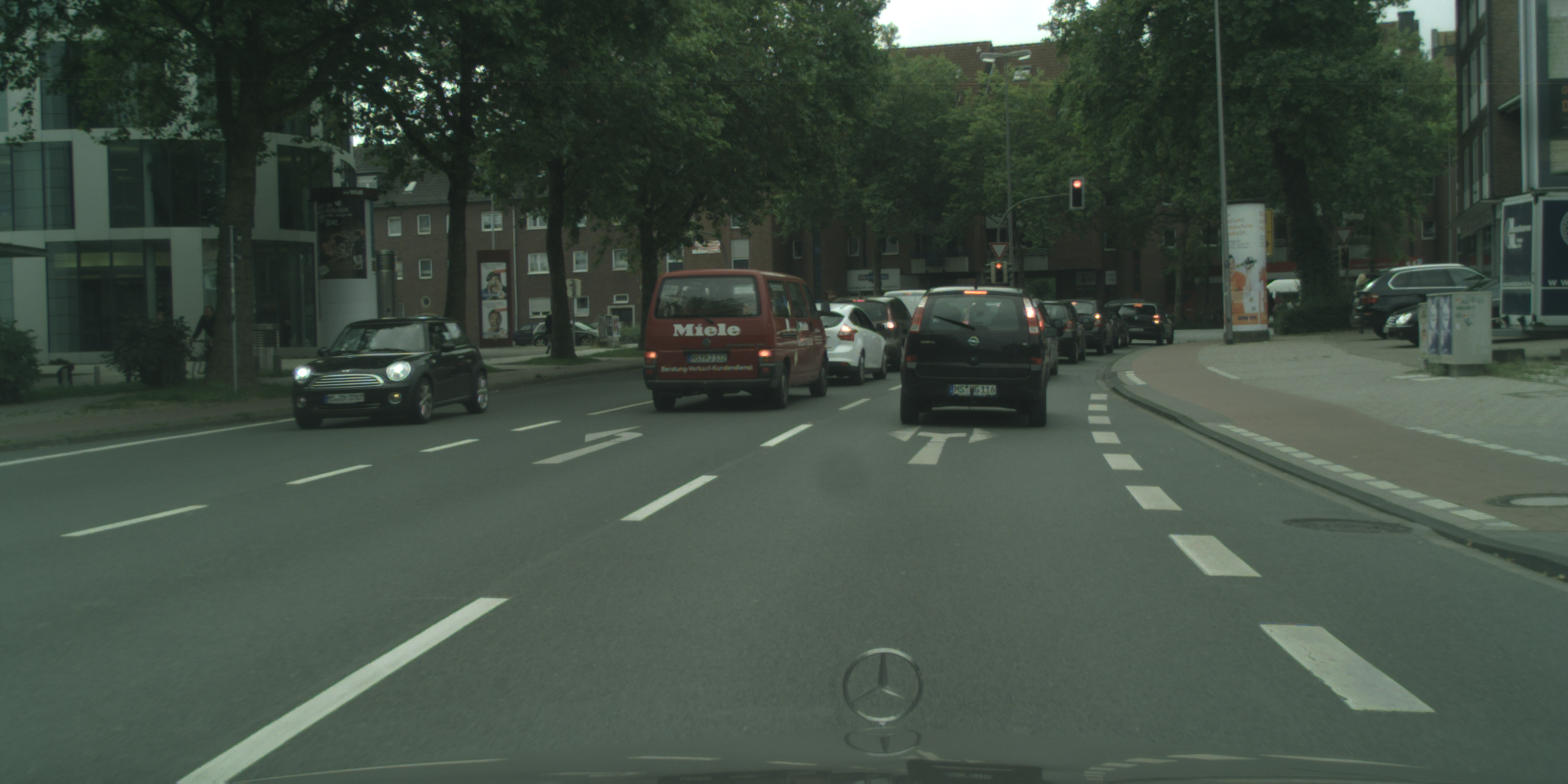}\\
			\vspace{0.02\textwidth}
			\includegraphics[width=1\textwidth]{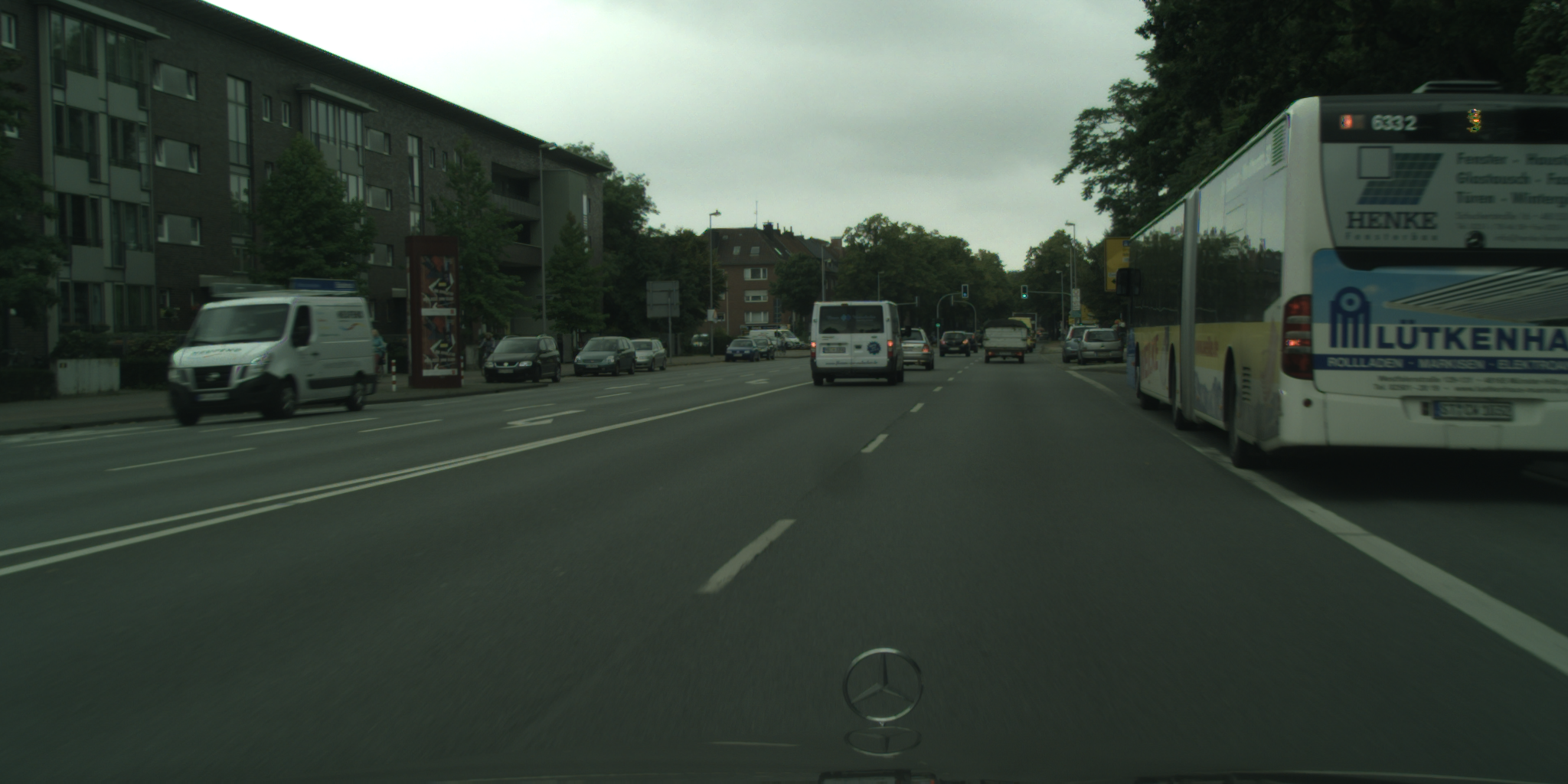}\\
			\vspace{0.02\textwidth} 
			\includegraphics[width=1\textwidth]{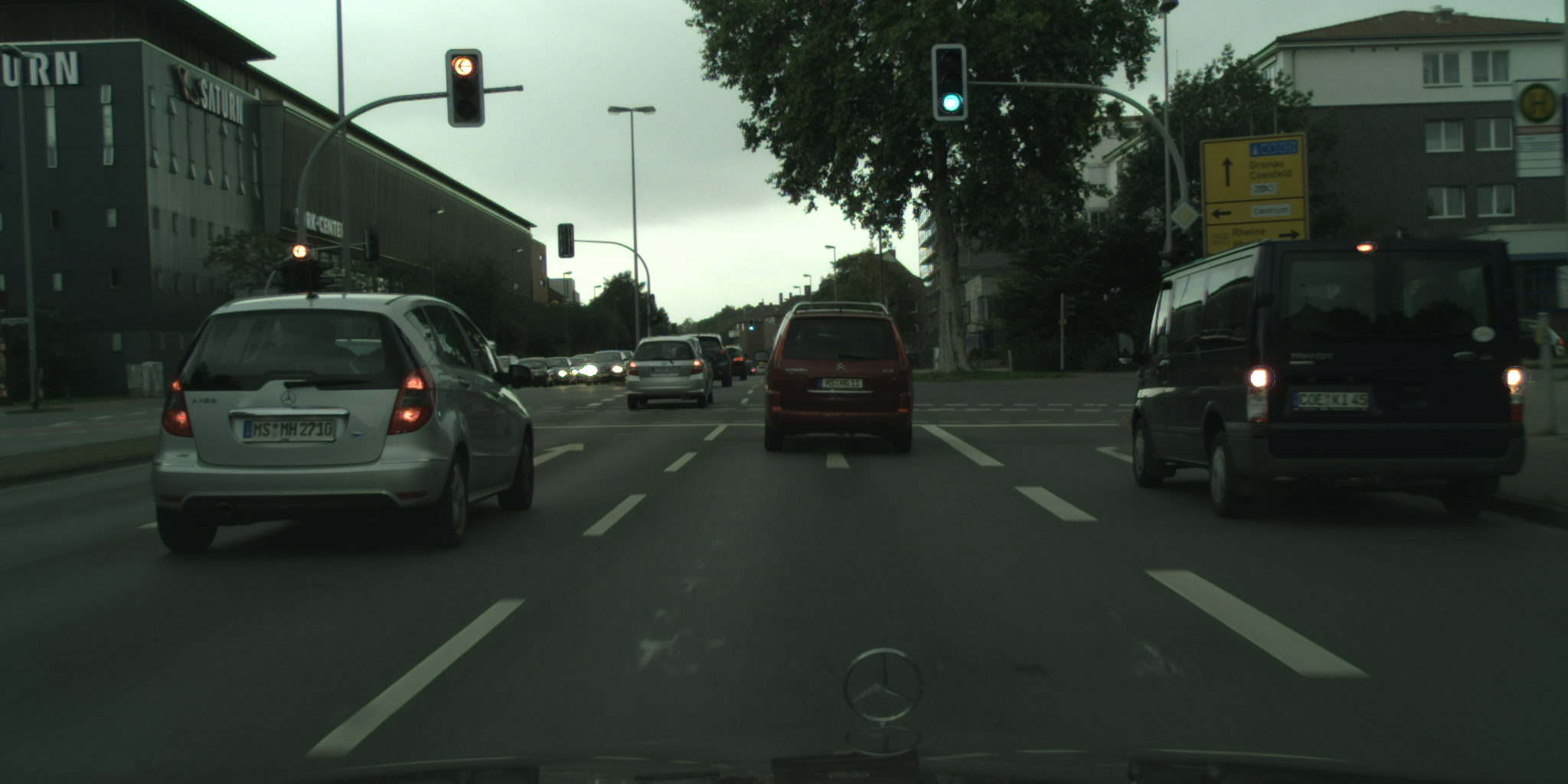}\\
			\vspace{0.2cm} 
			\label{image}	
		\end{minipage}
	}\hspace{-0.2cm}
	\subfigure[Groundtruth]{
	\begin{minipage}[t]{0.19\textwidth}
		\centering
		\includegraphics[width=1\textwidth]{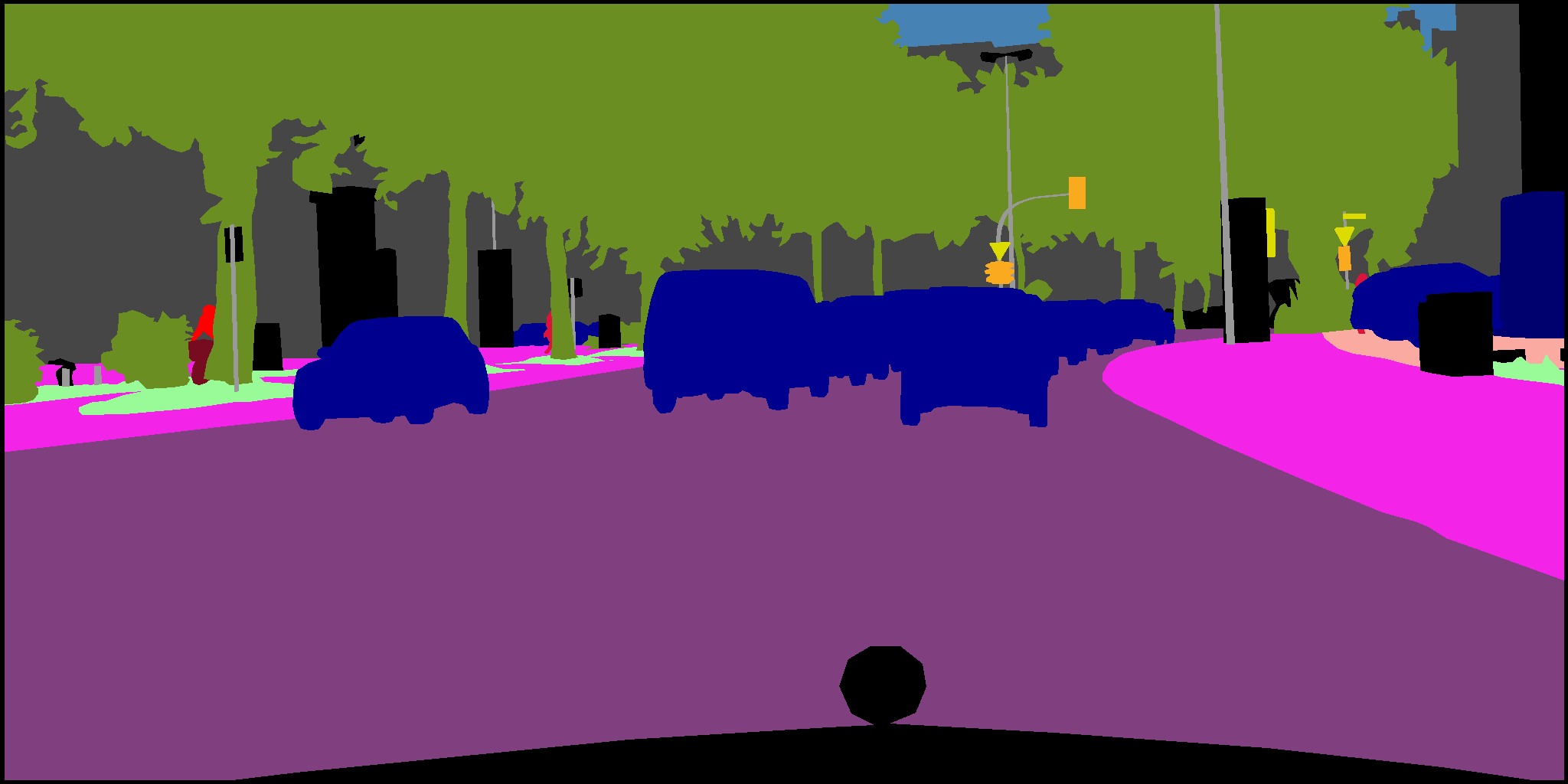}\\
		\vspace{0.02\textwidth}
		\includegraphics[width=1\textwidth]{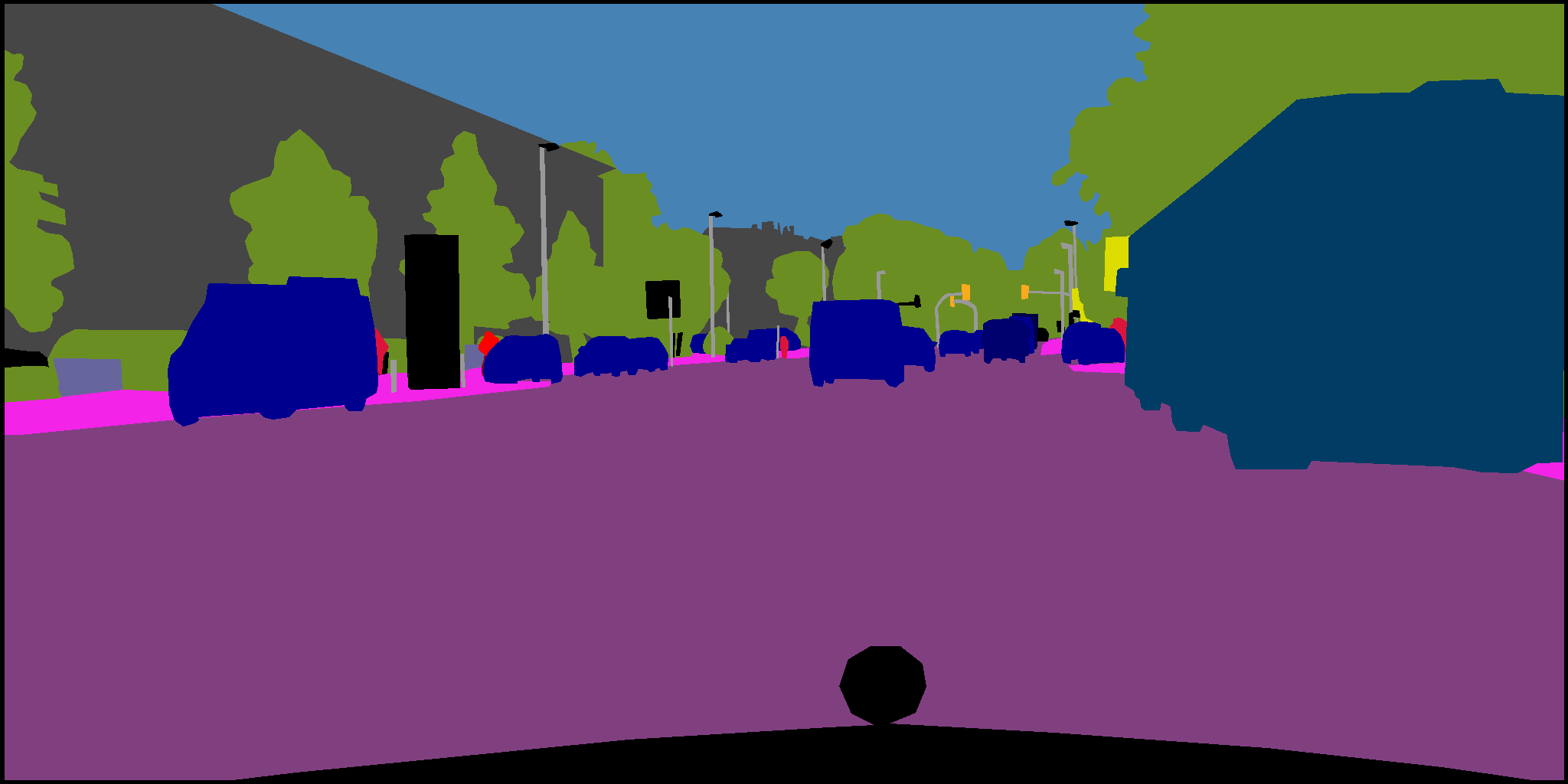}\\
		\vspace{0.02\textwidth} 
		\includegraphics[width=1\textwidth]{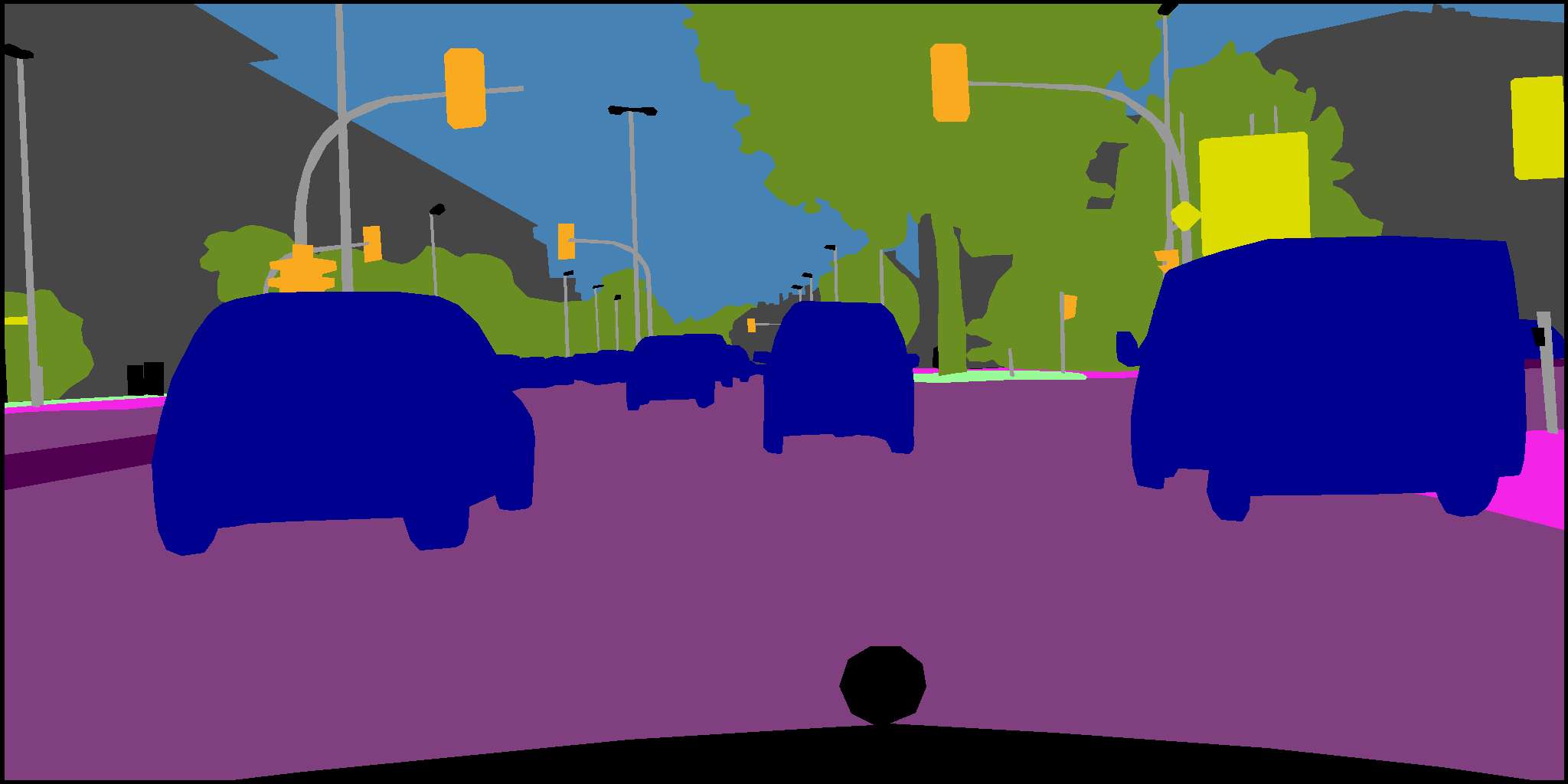}\\
		\vspace{0.2cm} 
		\label{gt}
	\end{minipage}
	}\hspace{-0.2cm}
	\subfigure[$768\times384$]{
		\begin{minipage}[t]{0.19\textwidth}
			\centering
			\includegraphics[width=1\textwidth]{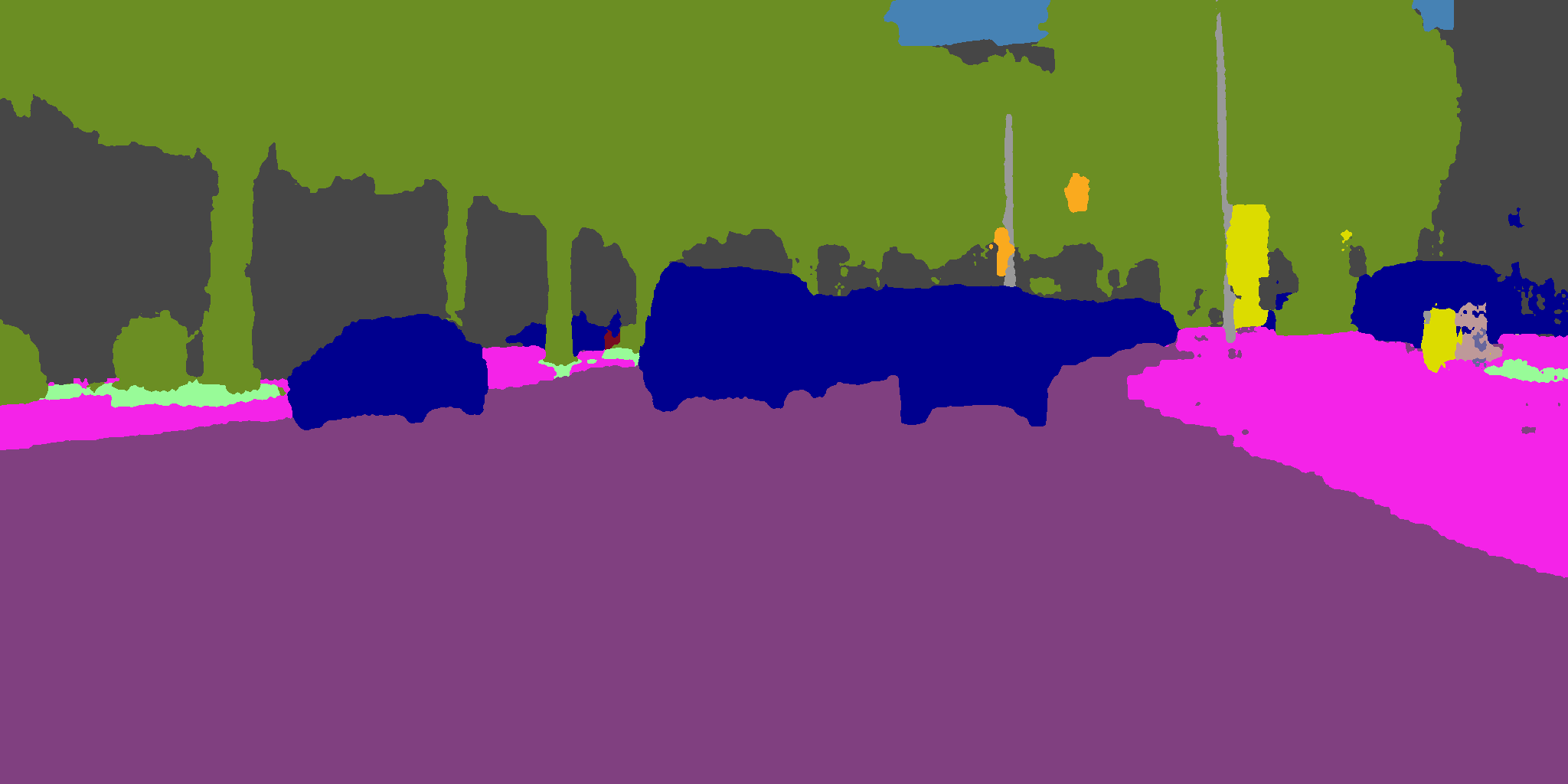}\\
			\vspace{0.02\textwidth}
			\includegraphics[width=1\textwidth]{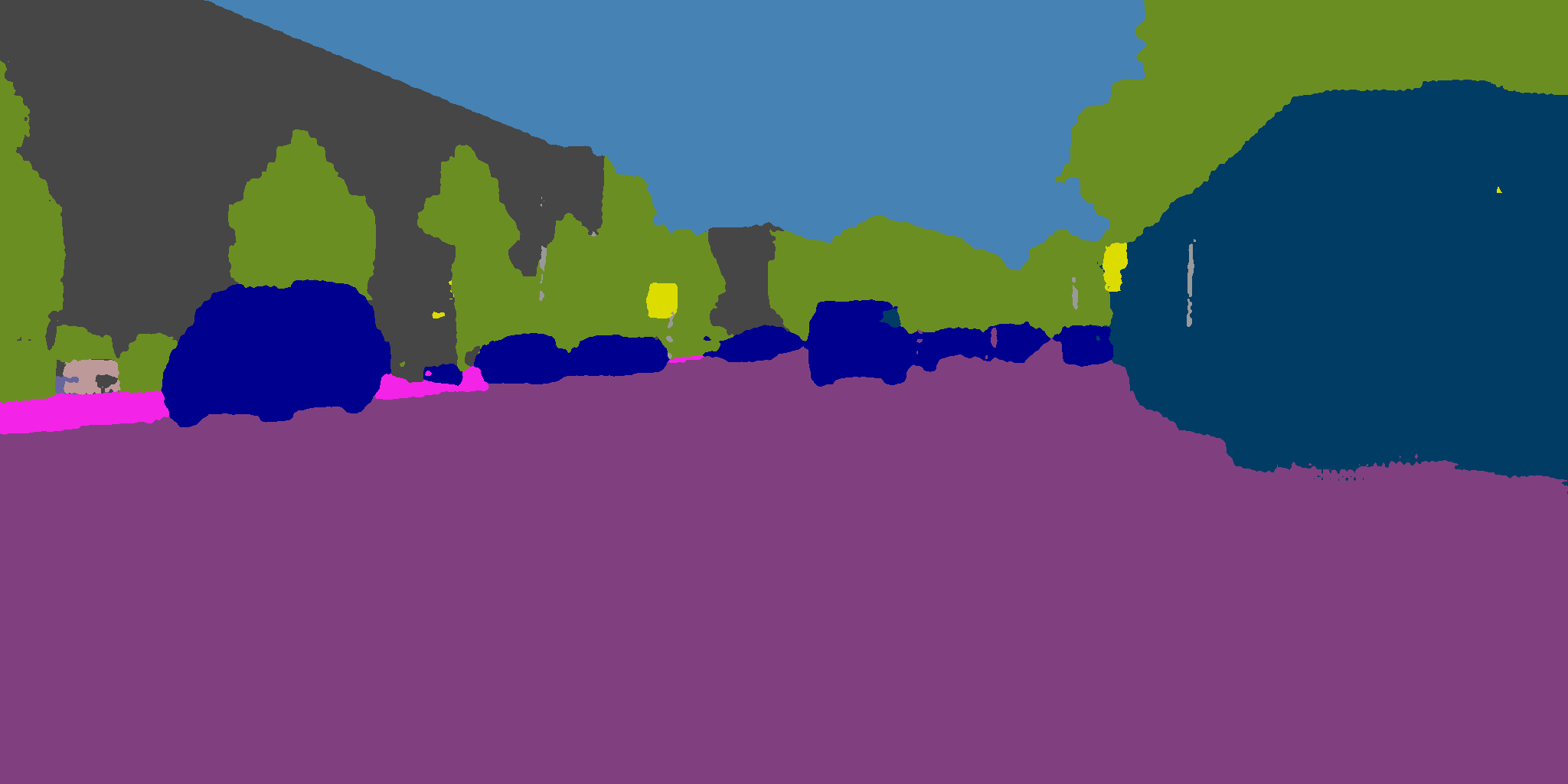}\\
			\vspace{0.02\textwidth} 
			\includegraphics[width=1\textwidth]{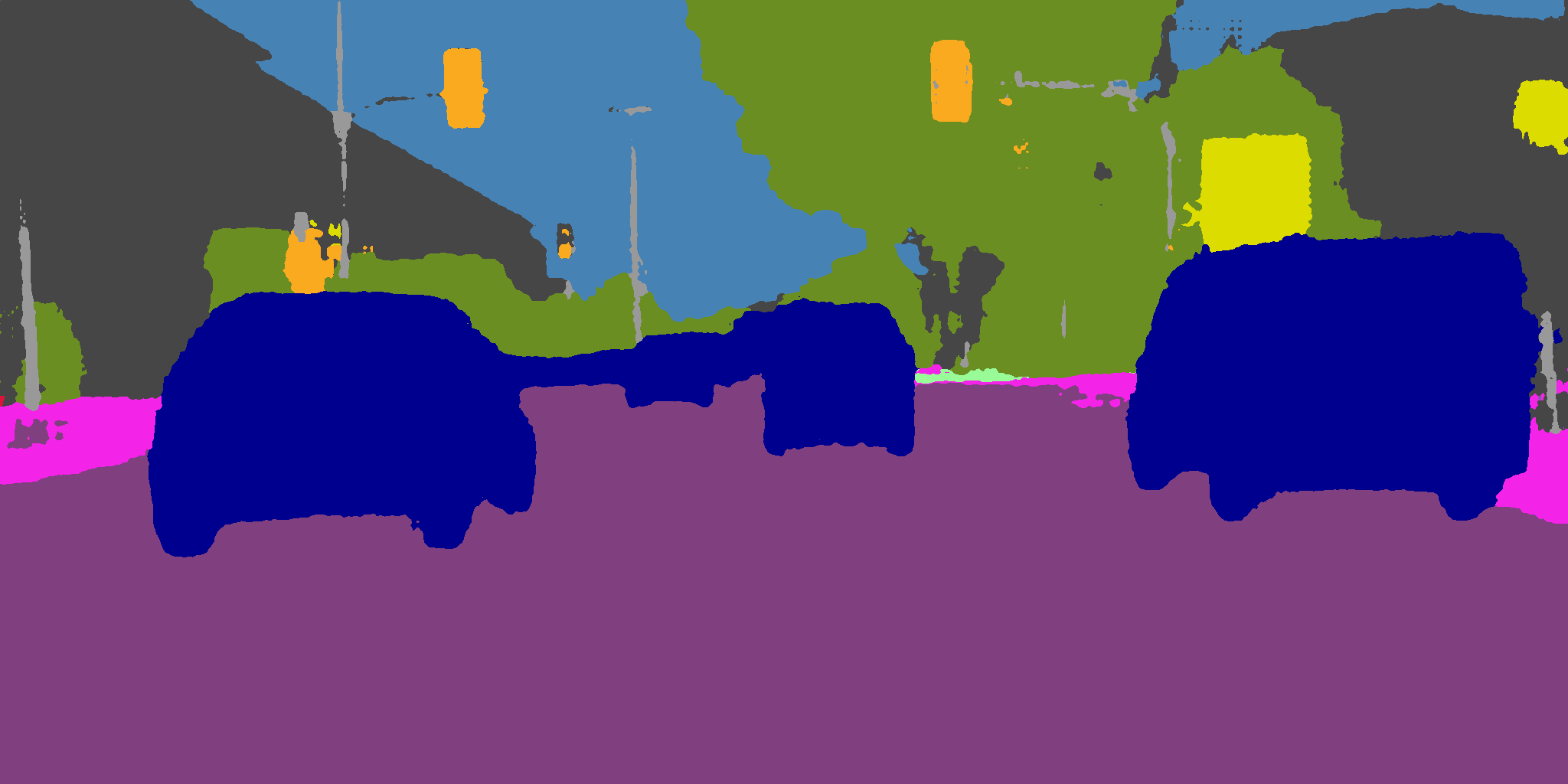}\\
			\vspace{0.2cm} 
			\label{canet}
		\end{minipage}
	}\hspace{-0.2cm}	
	\subfigure[$1024\times512$]{
	\begin{minipage}[t]{0.19\textwidth}
		\centering
		\includegraphics[width=1\textwidth]{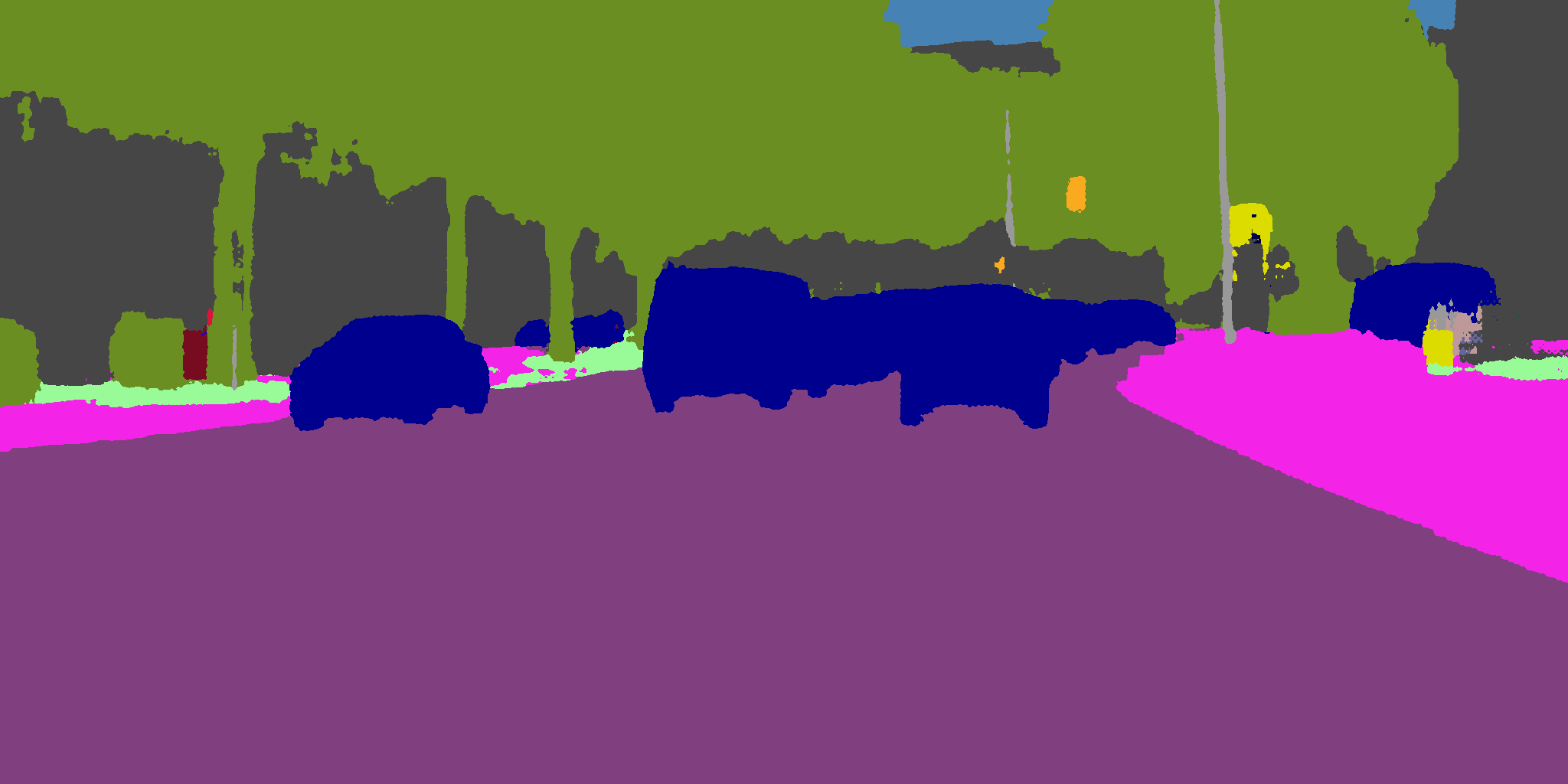}\\
		\vspace{0.02\textwidth}
		\includegraphics[width=1\textwidth]{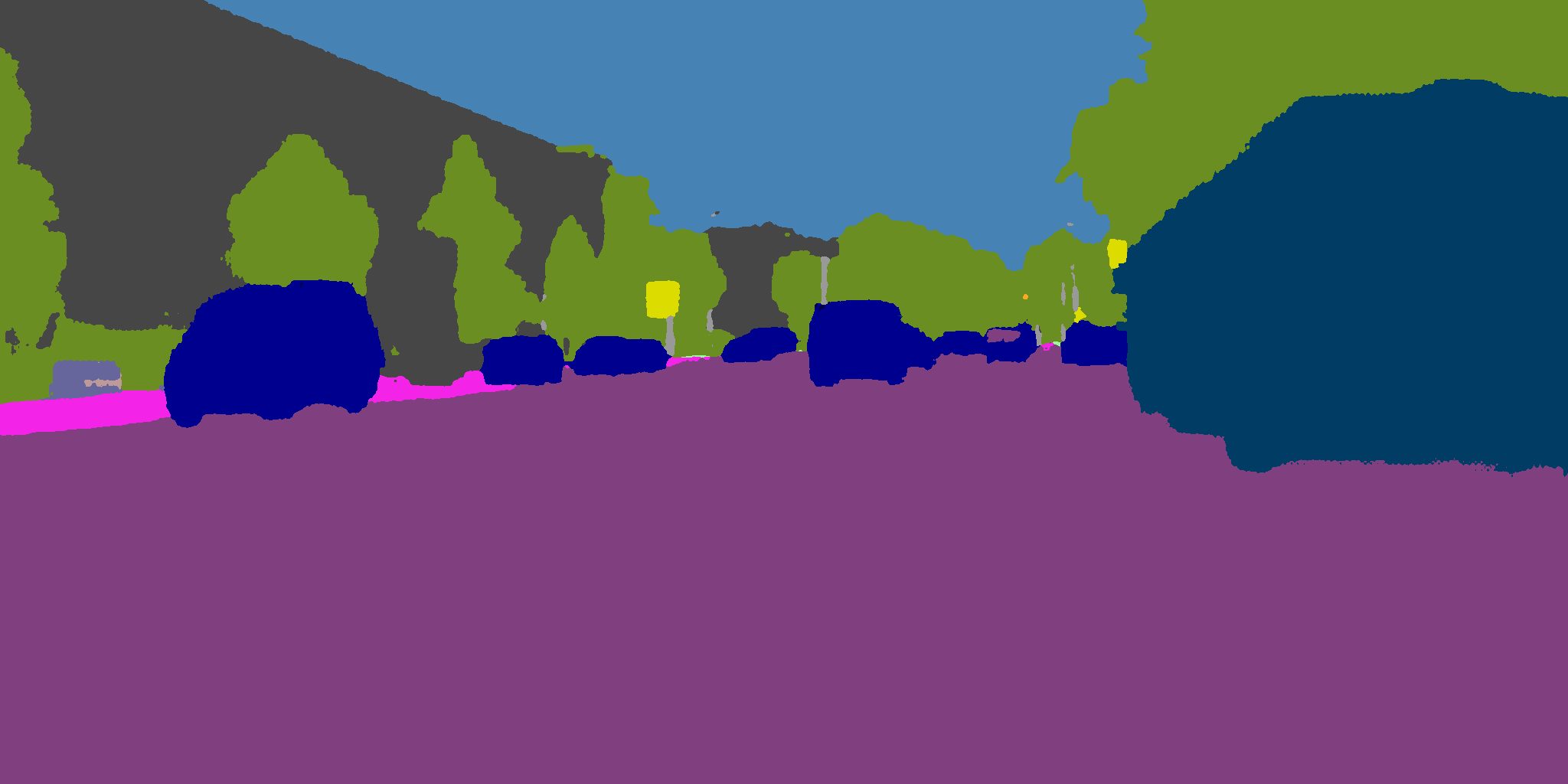}\\
		\vspace{0.02\textwidth} 
		\includegraphics[width=1\textwidth]{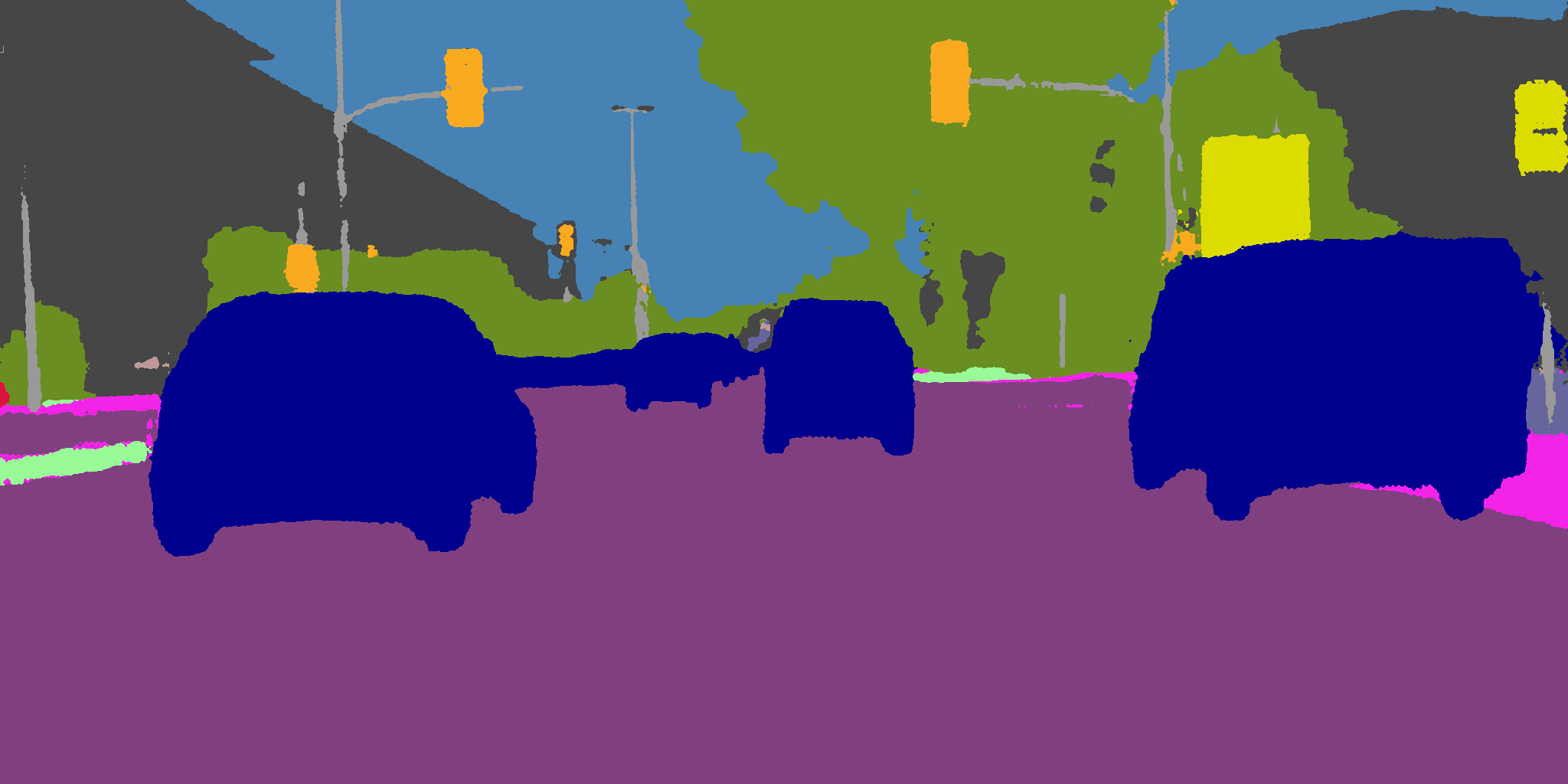}\\
		\vspace{0.2cm} 
		\label{canet}
	\end{minipage}
	}\hspace{-0.2cm}	
	\subfigure[$1536\times768$]{
	\begin{minipage}[t]{0.19\textwidth}
		\centering
		\includegraphics[width=1\textwidth]{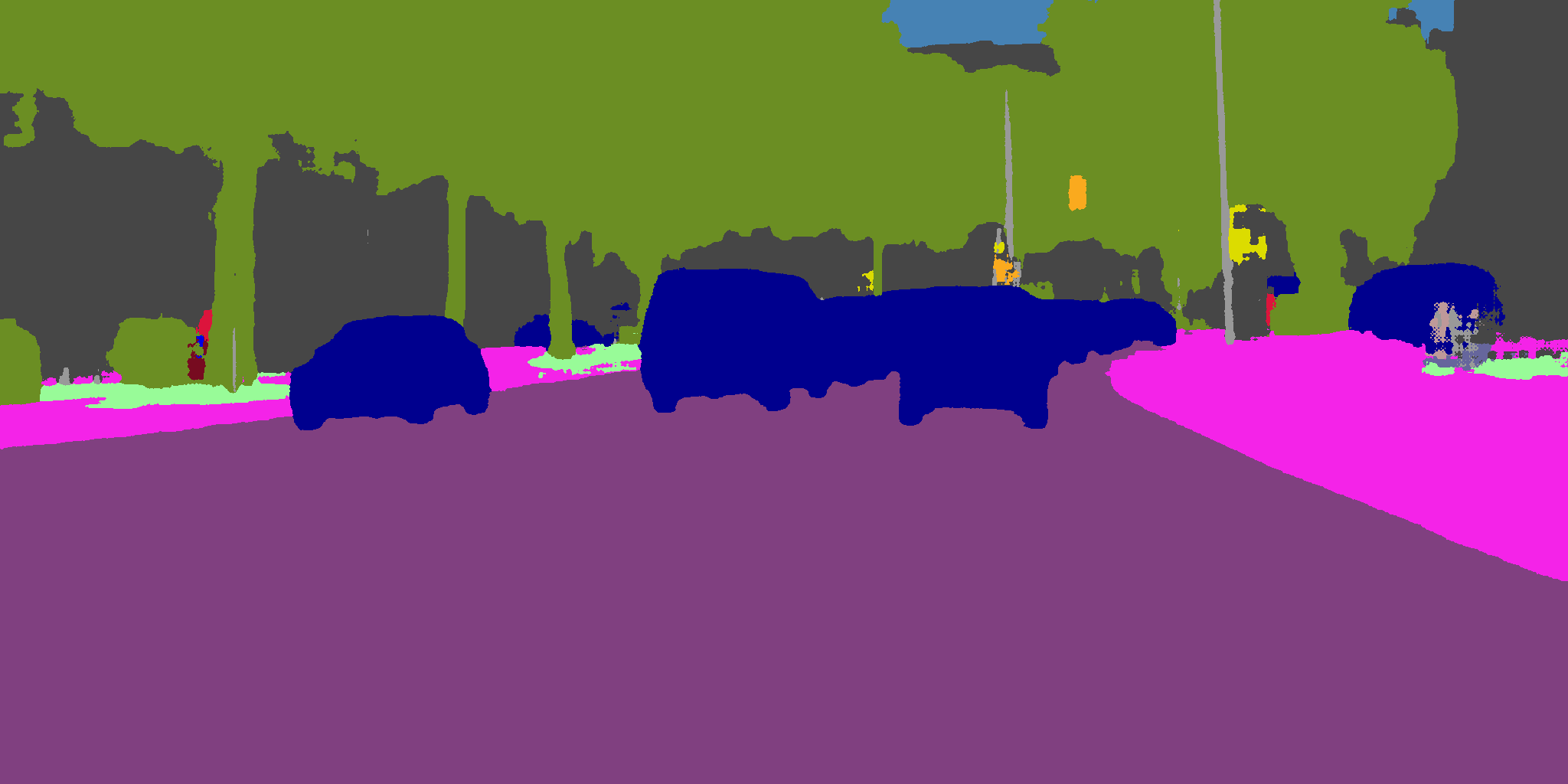}\\
		\vspace{0.02\textwidth}
		\includegraphics[width=1\textwidth]{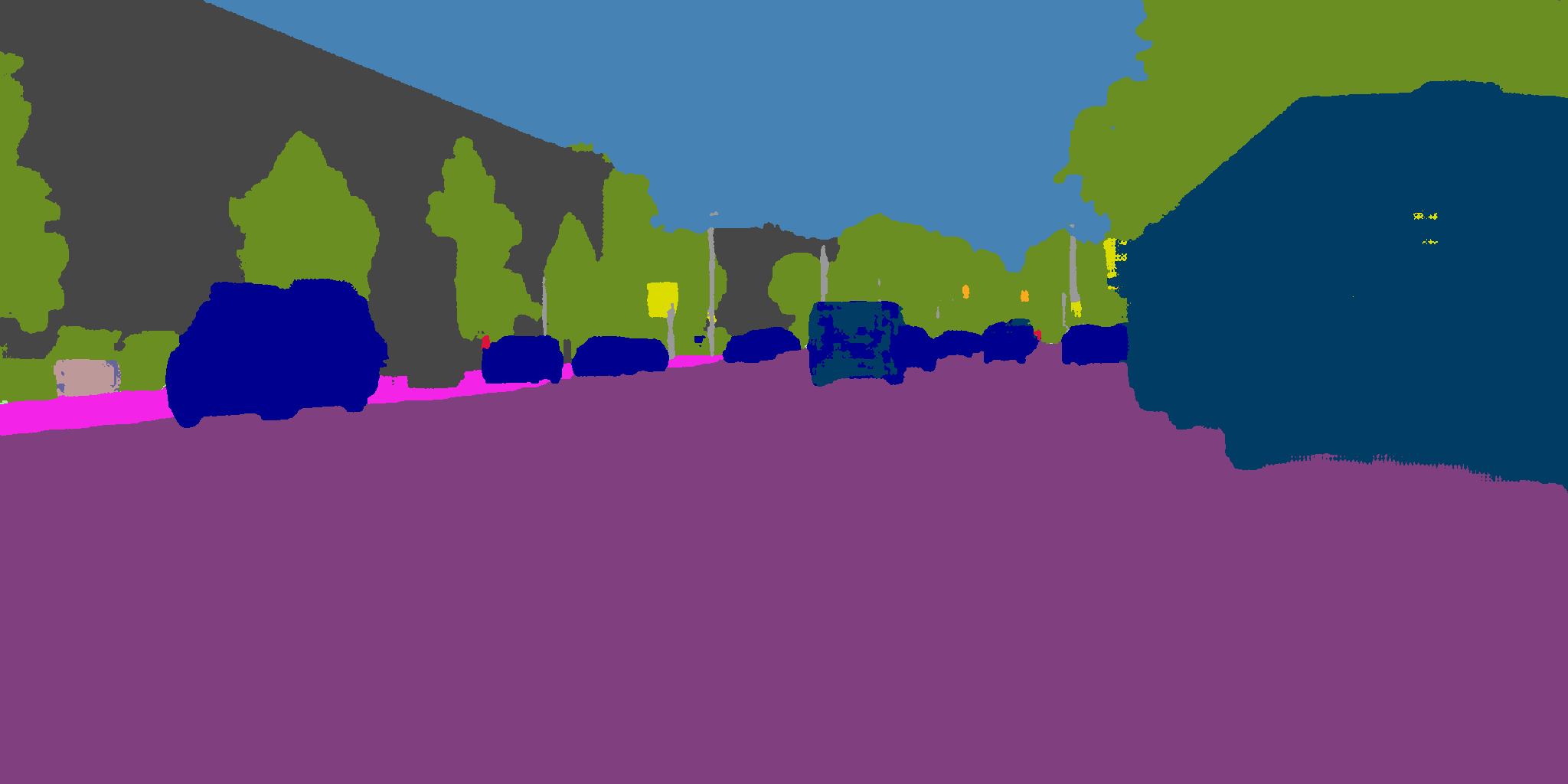}\\
		\vspace{0.02\textwidth} 
		\includegraphics[width=1\textwidth]{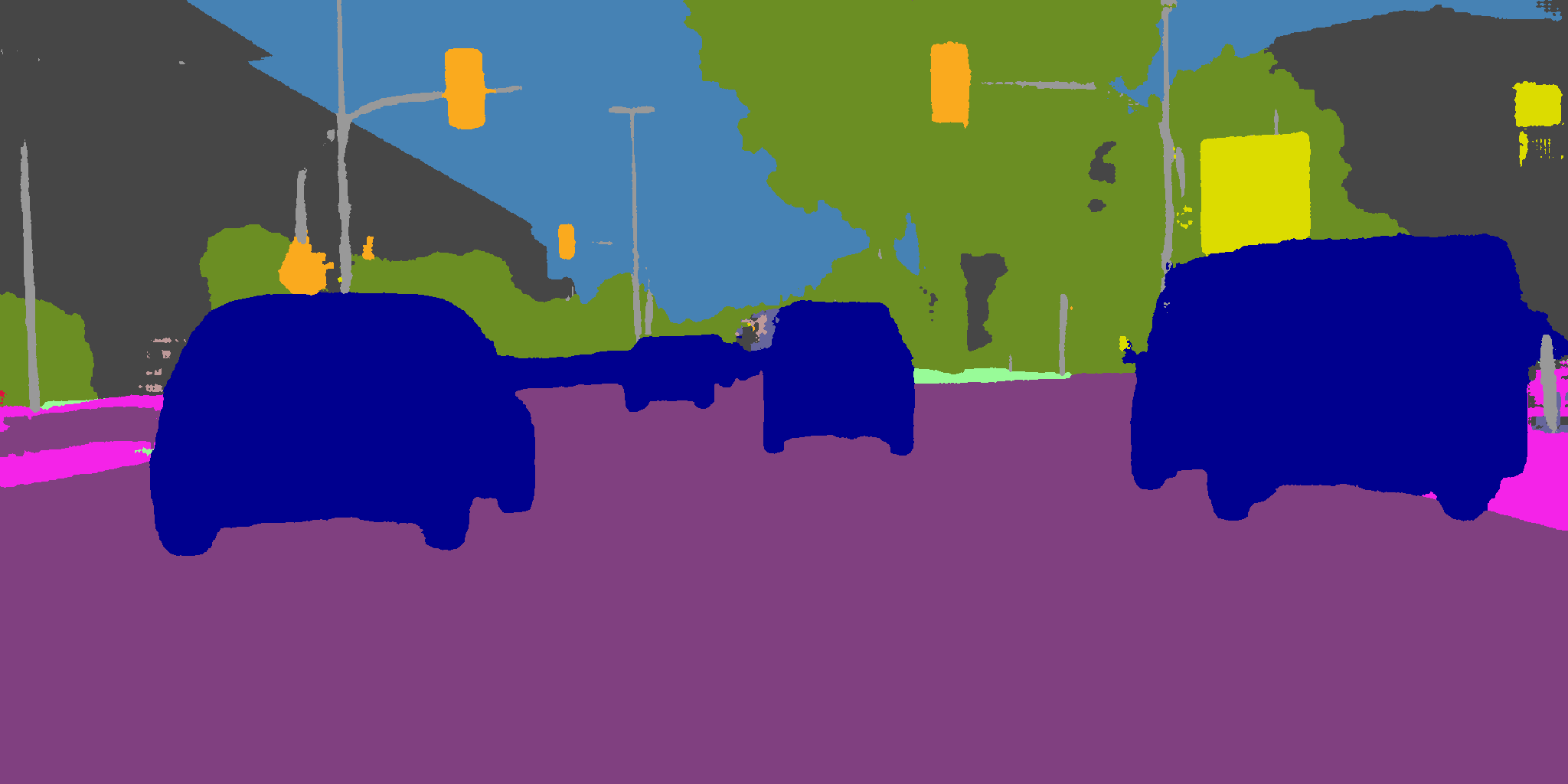}\\
		\vspace{0.2cm} 
		\label{canet}
	\end{minipage}
	}	
	\caption{Visualization results on Cityscapes dataset with $768\times384$, $1024\times512$ and $1536\times768$ input resolutions.}
	\label{example}
\end{figure}

As shown in Table \ref{city}, the number of parameters of FPENet is close to the ESPNet, but the accuracy is 7.7\% higher at the same input size. FPENet is 14 and 19 times smaller than the BiSeNet1 and ICNet, while the mIoU is only 0.4\% and 1.5\% less, respectively. Besides, FPENet achieves 102 FPS speed at $1024\times512$ input resolution, which significantly outperforms most of existing real-time methods. When the input size is $768\times384$, the accuracy is still better than some methods with lower FLOPs. To improve the accuracy, we also used $1536\times768$ resolution for training and test. Some segmentation results of FPENet with different input resolutions are presented in Fig. \ref{example}.

\begin{table}[t!]
	\begin{center}
		\begin{tabular}{ l  c  c  c c c}
			\hline
			Method                           & Input size       & \#Params & FLOPs & FPS & mIoU (\%) \\ \hline\hline
			ENet\cite{paszke2016}            & $1024\times512$  & 0.4M     & 4.4G  & 61  & 58.3      \\
			ESPNet\cite{mehta2018}           & $1024\times512$  & 0.4M     & 4.7G  & 132 & 60.3      \\
			ESPNetv2\cite{mehta2019espnetv2} & $1024\times512$  & 0.7M     & 3.5G  & 84  & 62.1      \\
			CGNet\cite{wu2018}               & $2048\times1024$ & 0.5M     & 28.0G & 14  & 64.8      \\
			ContextNet\cite{poudel2018}      & $2048\times1024$ & 0.9M     & 48.3G & 24  & 66.1      \\
			BiSeNet1\cite{yu2018}            & $1536\times768$  & 5.8M     & 14.8G & 79  & 68.4      \\
			ICNet\cite{zhao2018icnet}        & $2048\times1024$ & 7.8M     & 29.8G & 59  & 69.5      \\ \hline
			FPENet                           & $768\times384$   & 0.4M     & 3.2G  & 129 & 62.7      \\
			FPENet                           & $1024\times512$  & 0.4M     & 5.7G  & 102 & 68.0      \\
			FPENet                           & $1536\times768$  & 0.4M     & 12.8G & 55  & 70.1      \\ \hline
		\end{tabular}
	\end{center}
	\caption{Speed and accuracy comparison of FPENet on Cityscapes test set.}
	\label{city}
\end{table}

\subsection{CamVid}
The CamVid road scenes dataset has fully labelled images for semantic segmentation: 367 for training, 101 for validation and 233 for test. Each image is of $480\times360$ pixels, labelled with 11 semantic classes. We used the training and validation set to train our model and tested on the test set. Results of global accuracy and mIoU are shown in Table \ref{camvid}. Our method outperforms other deep models with fewer parameters.

\begin{table}[htb]
	\begin{center}
		\begin{tabular}{ l  c c  c}
			\hline
			Method                                     & \#Params & Global avg. (\%) & mIoU (\%) \\ \hline\hline
			ENet\cite{paszke2016}                      & 0.4M     & ---              & 51.3      \\
			FCN8 \cite{long2015}                       & 134.5M   & 83.1             & 52.0      \\
			Bayesian SegNet \cite{kendall2015bayesian} & 29.5M    & 86.9             & 63.1      \\
			BiSeNet1\cite{yu2018}                      & 5.8M     & ---              & 65.6      \\ \hline
			FPENet                                     & 0.4M     & 89.6             & 65.4      \\ \hline
		\end{tabular}
	\end{center}
	\caption{Results on CamVid test set. ``---'' indicates that the methods do not report the corresponding results. }
	\label{camvid}
\end{table} 

\section{Conclusions}
This paper presents a lightweight architecture, feature pyramid encoding network (FPENet) for semantic segmentation. A feature pyramid encoding (FPE) block is proposed and adopted in every stage of FPENet to encode multi-scale features using a spatial pyramid of depthwise dilated convolutions. Mutual embedding upsample (MEU) modules are employed in the decoder to aggregate features from different stages. The ablation experiments show that FPE blocks significantly improve accuracy due to large receptive field and enhanced information flow, and the MEU modules aggregate deep contextual features and shallow spatial features efficiently. Experimental results on the Cityscapes and CamVid datasets demonstrate superiority of the purposed FPENet over other real-time methods with much faster inference speed and fewer parameters.     

\bibliography{refs}
\end{document}